\pdfoutput=1

\documentclass[11pt]{article}

\usepackage[]{ACL2023}

\usepackage{times}
\usepackage{latexsym}
\usepackage{amsfonts}
\usepackage{url}
\usepackage{graphicx}
\usepackage{amsmath}
\usepackage{multirow}
\usepackage[raggedrightboxes]{ragged2e}
\usepackage{booktabs}
\usepackage[utf8]{inputenc}

\usepackage{microtype}

\usepackage{inconsolata}

\DeclareMathOperator*{\argmax}{arg\,max}

\usepackage[T1]{fontenc}

\usepackage[utf8]{inputenc}

\usepackage{microtype}

\usepackage{inconsolata}

\title{Towards Argument-Aware Abstractive Summarization of Long Legal Opinions with Summary Reranking}

\author{Mohamed Elaraby, Yang Zhong, Diane Litman \\
        University of Pittsburgh \\ Pittsburgh, PA, USA \\ \texttt{\{mse30,yaz118,dlitman\}@pitt.edu}}

\begin{document}
\maketitle
\begin{abstract}

We  propose a simple approach for the abstractive summarization of long legal opinions that considers the argument structure of the document. Legal opinions often contain  complex and nuanced argumentation, making it challenging to generate a concise summary that accurately captures the main points of the legal opinion. Our approach involves 
using argument role information to generate multiple candidate summaries, 
then reranking these candidates based on 
alignment with the document's argument structure.
We demonstrate the effectiveness of our approach on a dataset of 
long legal opinions and show that it outperforms several 
strong baselines. 

\end{abstract}

\section{Introduction}

Legal opinions contain implicit argument structure spreading across long texts. 
Existing summarization models often struggle to accurately capture the main arguments of such documents, leading to summaries that are suboptimal \cite{xu2021toward, elaraby-litman-2022-arglegalsumm}.
We propose an 
approach for the abstractive summarization of long legal opinions that leverages argument structure. 

Legal opinions often follow a specific argumentative structure, with the main points of the argument being presented clearly and logically \cite{xu2021toward,habernal2022mining, xu2022multi}. Prior work has shown that by considering this structure during summarization, it is possible to generate extractive and abstractive summaries that more accurately reflect the original argumentation in the document \cite{elaraby-litman-2022-arglegalsumm, zhong2022computing,agarwal2022extractive}. In this paper, we present a framework for abstractive summarization of long legal opinions that extends this literature by {\it leveraging argument structure during summary reranking} to both generate and score candidates. 
Our method involves utilizing the Longformer-Encoder-Decoder (LED) \cite{beltagy2020longformer} model to generate multiple candidate summaries by training it on various input formats. This allows for the consideration of different argument representations in the summary generation process. Additionally, we use beam search to further diversify the output. Finally, we rank the candidate summaries by measuring their lexical similarity to the input's main arguments. 


We evaluate our approach on a dataset of long legal opinions obtained from the Canadian Legal Information Institute (CanLII)\footnote{Data was obtained through an agreement with CanLII (\url{https://www.canlii.org/en/}).} and demonstrate 
that our method outperforms 
competitive baselines. 
Our results with ROUGE and BERTScore \cite{lin2004rouge, zhang2019bertscore} suggest that considering the argumentative coverage of the original opinions  can lead to a more effective selection of summaries.

Our contributions 
are: \textbf{(1)} We propose a simple 
reranking approach that takes into account the argumentative structure of legal opinions to improve over the standard finetuning of generation models. \textbf{(2)} We demonstrate through empirical results and ablation analysis reasons for the effectiveness of our approach for summarizing long legal opinions. Our code can be accessed through this repository: \url{https://github.com/EngSalem/legalSummReranking}

\section{Related Work}

\textbf{Long Legal Document Summarization}
Legal documents have a distinct format, with a hierarchical structure and specialized vocabulary that differs from that of other domains  \cite{kanapala2019text}. They also tend to be longer in length \cite{kan-etal-2021-home, huang2020legal,moro2022semantic},
which has led to the use of transformer models with sparse attention mechanisms \cite{michalopoulos-etal-2022-icdbigbird, guo-etal-2022-longt5, beltagy2020longformer} 
to reduce the complexity of encoding lengthy text. Legal {\it opinions}, in particular, have a complex argumentative structure that spans across the text, making it crucial to address in summaries \cite{xu2021toward, xu2022multi, elaraby-litman-2022-arglegalsumm}. {\it We use  prior legal opinion summarization methods as evaluation baselines}. 

\textbf{Summarization and Argument Mining} 
 Using a dialogue summarization dataset with argument information, \citet{fabbri2021convosumm} converted an argument graph into a textual format to train a summarizer. For legal documents, \citet{agarwal2022extractive} used argument role labeling to improve extractive summarization using multitask learning. \citet{elaraby-litman-2022-arglegalsumm} blended argument role labeling and abstractive summarization using special markers, generating summaries that better aligned with legal argumentation. {\it We incorporate the models of \citet{elaraby-litman-2022-arglegalsumm} into summary reranking and further improve
 performance.}

\textbf{Second Stage Reranking}
Generating multiple outputs and reranking them according to certain criteria has been successfully applied in NLP downstream applications including abstractive summarization. Some methods use different input formats to generate multiple outputs. \citet{oved2021pass} perturbed input multi-opinion reviews to generate multiple candidate summaries, then ranked them using coherency.  \citet{ravaut2022summareranker} used a multitask mixture of experts to directly model the probability that a summary candidate is the best one.  
\citet{liu2021simcls} ranked candidate summaries generated from 16 diverse beam searches to improve news summarization in terms of ROUGE score.  \citet{liu2022brio} presented a novel technique for summary reranking that involves a non-deterministic training objective. Their approach enables the model to directly rank the summaries that are probable from beam-search decoding according  to their quality.
{\it 
We rely on distinct argument-aware input formats in addition to diverse beam decoding to 
develop our argument-aware 
reranking method}.

\begin{figure*}[h]
\small
\begin{center}

 \includegraphics[width=14cm,height=7cm]{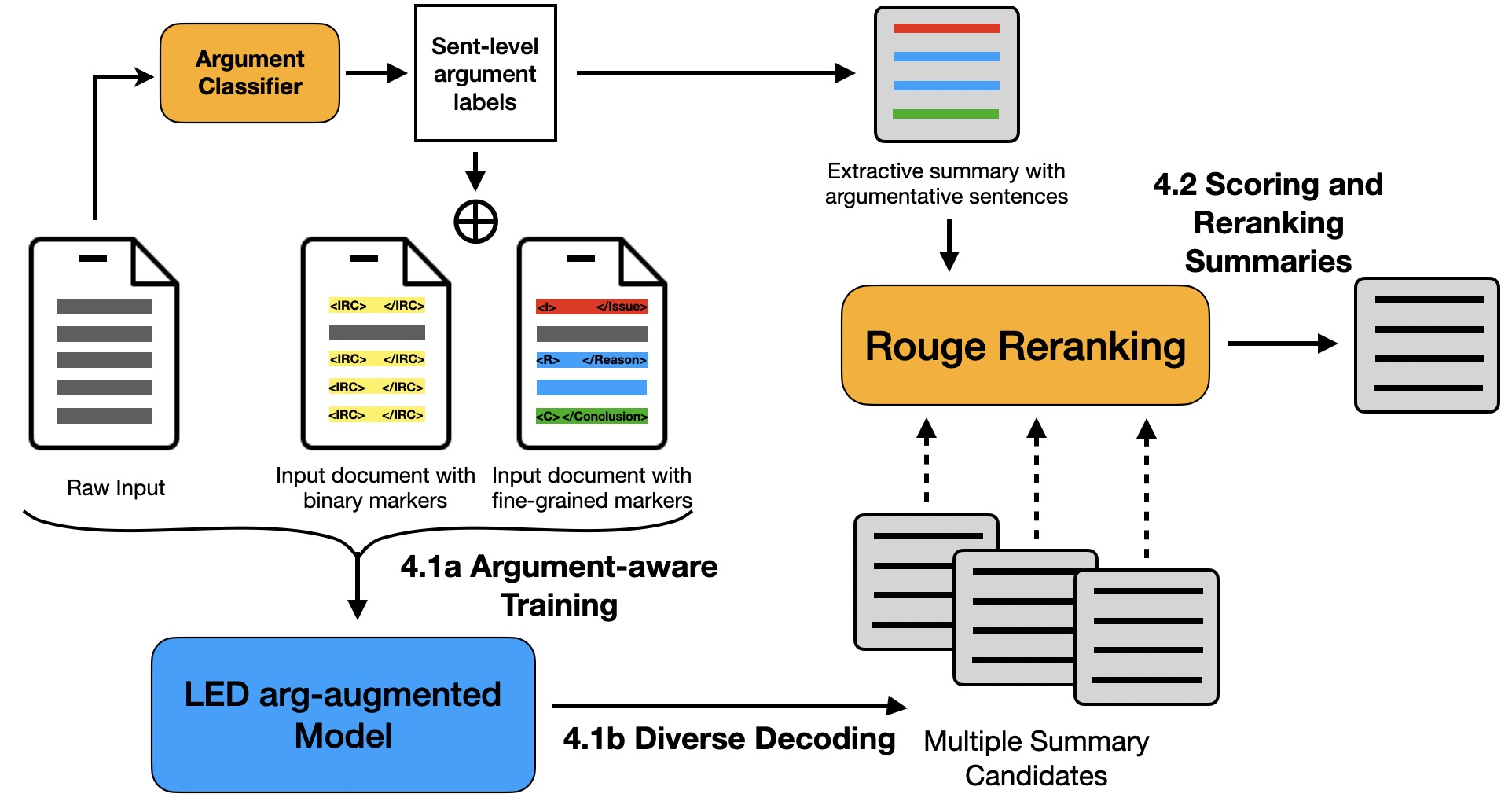}
 \caption{\label{sys_diag} Illustration of basic components of our approach. For input documents with fine-grained markers, colored sticks are sentences with argument role labels of \textcolor{red}{Issue}, \textcolor{blue}{Reason}, and \textcolor{green}{Conclusion}. We used one \textcolor{yellow}{IRC} label for the binary version. In our real dataset, marked sentences are surrounded with special markers (Appendix  \ref{markers}).}
 \end{center}
\end{figure*}

\section{Annotated Dataset}

 \makeatletter
\def\hlinewd#1{%
\noalign{\ifnum0=`}\fi\hrule \@height #1 \futurelet
\reserved@a\@xhline}
\makeatother
    

We employ the annotated subset \cite{xu2021toward, elaraby-litman-2022-arglegalsumm} of the \textbf{CanLII} dataset \cite{zhong2022computing} used in prior 
summarization research of legal opinions.
This subset contains $1049$ 
opinion/summary pairs 
annotated with sentence-level argument role labels for both input documents and reference summaries. 
The input 
opinions have mean/max lengths of 4375/62786 words, motivating us to use models for long text.




Recent work has proposed  argument role taxonomies aligned with structures commonly found in legal text \cite{habernal2022mining, xu2021toward}. The CanLII data was {\bf annotated for argument roles} using the {\bf IRC scheme} for legal opinions \cite{xu2021toward}, 
which divides argument roles into \textbf{Issues} (legal questions which a court addressed in the document), \textbf{Reasons} (pieces of text which indicate why the court reached the specific conclusions), and  \textbf{Conclusions} (court’s decisions for the corresponding issues).   We use these 3 fine-grained IRC labels, as well as collapse them into a single argumentative label, to incorporate argument structure into our models. 
An IRC-annotated opinion and summary pair  can be found in Appendix \ref{irc_des}.

\section{Model and Methods}
\label{sec:models}

Our proposed method follows the generate and ranking paradigm and can be split into two parts. First, we  explore techniques to utilize an argumentation augmented LED model
to generate multiple candidate summaries  $\mathbb{S}$.   Second, we propose a function $\mu$ that scores a summary $S$ where $S \in \mathbb{S}$ based on its argumentative alignment with the input document.  The best candidate $S^*$ is selected such that $S^* = \argmax_{S_i \in \mathbb{S}} \{\mu(S_1), \mu(S_2),.., \mu(S_n) \}$.
Figure \ref{sys_diag} shows an overview of our approach. 

\subsection{Generating Candidates: Argument-Aware Training  + Diverse Decoding}

Diverse decoding techniques such as  beam-search can help diversify the {\it summary output}; however, it's only limited to the underlying language model used in the decoder and is completely isolated from the input format. 
Alternatively, we propose to complement the beam search via finetuning  LED 
on three {\it different input formats}. We refer to  this model as $M_{arg-augmented}$ such that the model parameter $\theta^*_{arg-augmented}$ is selected such that $$\theta^*_{arg-augmented} = \argmax_{\theta}P(S|\mathbb{X})$$ During finetuning, $S$ is the reference summary, $\theta$ represents the trainable model parameters, and $\mathbb{X}$ is a set of inputs $\mathbb{X} = \{X_{raw}, X_{arg\_binary} , X_{arg\_finegrained}\}$, where $X_{raw}$ is the input without the argument markers, $X_{arg\_binary}$ is the input document with binary argument markers added to highlight argument role sentences, and $X_{arg\_finegrained}$   is the input document with the fine-grained argumentative markers added to also delineate the roles  (i.e., Issue, Reason, Conclusion). These three representations of the input share the same  reference summary, meaning that we augmented the training data three times. Table \ref{example} shows an example of the distinct representations of our new training data.
At inference time, we use the predicted markers by adopting the argument mining code\footnote{We retrain the model (details in Appendix \ref{exp_setup}), yielding  a macro-average of 0.706 F1 on the four-way classification (Issues, Reasons, Conclusions,  Non-argumentative).} 
from \citet{elaraby-litman-2022-arglegalsumm} 
instead of the manually labeled ones to construct $\hat{X}_{arg\_binary}, \hat{X}_{arg\_finegrained}$ of $\hat{\mathbb{X}}$  where  $\hat{\mathbb{X}} = \{X_{raw}, \hat{X}_{arg\_binary}, \hat{X}_{arg\_finegrained} \}$. Our incentive is that different formats of the input would yield different generated summaries that take into account different representations of the argumentative structure in the input.



\begin{table}[ht]
\small
\begin{tabular}{p{0.15\textwidth}| p{0.27\textwidth}}
\hlinewd{1.5pt} 

\textbf{Input format} &  \textbf{Example} \\ \hlinewd{1.5pt}

\multirow{2}{*}{${X_{raw}}$} & $S_1 | S_2 | ... |$ Issue Sentence $|$ Reason Sentence $| ...$  \\ \hline

\multirow{3}{*}{${{X}_{arg\_binary}}$} & $S_1 | S_2 | ... |$ \texttt{<IRC>} Issue Sentence \texttt{</IRC>} $|$ \texttt{<IRC>} Reason Sentence \texttt{</IRC>} $| ...$   \\ \hline

\multirow{3}{*}{${{X}_{arg\_finegrained}}$} & $S_1 | S_2 | ... |$ \texttt{<Issue>} Issue Sentence \texttt{</Issue>} $|$ \texttt{<Reason>} Reason Sentence \texttt{</Reason>} $| ...$ \\ \hlinewd{1.5pt}

\end{tabular}
\caption{An example of $\mathbb{X}$, which consists of three data points in different formats that share the same reference summary. In the table, $S1$ refers to the first sentence of the text, $S2$ to the second sentence, and so on. \texttt{<IRC>}, \texttt{<Issue>}, and \texttt{<Reason>} are the argumentative marker tokens described in Appendix \ref{markers}. \label{example}}
\end{table}

\begin{center}
    
\begin{table*}[]
\small
\renewcommand{\arraystretch}{1.15}
\begin{tabular}{p{0.155\textwidth}cp{0.32\textwidth}lllll}

\hlinewd{1.5pt}
\textbf{Experiments}          &    ID                          & \textbf{Model }                                              & \textbf{R-1 }                & \textbf{R-2}        & \textbf{R-L}                 & \textbf{BS}  & \textbf{src. marker} \\ \hlinewd{1.5pt}
\multirow{5}{*}{\textbf{Abstractive baselines}}           &1   & finetune LED-base                                   & 47.33                   & 22.80           & 44.12                   & 86.43           & - \\  
    \cline{2-8} 
    &2 & {arg-LED-base (binary markers)}                       & {48.85}                   & {24.74}          & {45.82}                   & {86.79}          & \multirow{2}{*}{{predicted }}                 \\  
                                                 &3 &  {arg-LED-base (fine-grained markers)}                 & {49.02}                   & {24.92}          & {45.92}                   & {86.86}         &                 \\ 
                                                    \cline{2-8}
   
     &4 & arg-LED-base (binary markers)                      & 50.64                   & 26.62          & 47.48                   & 86.90          & \multirow{2}{*}{oracle}            \\  
                                                  &5 & arg-LED-base (fine-grained markers)                 & 51.07                   & 27.06          & 48.01                   & 86.92         &                 \\

                                                   \hline
\multirow{4}{*}{\textbf{Ranking baselines}}            &6 & {baseline ranking}                           & {49.79}                   & {25.13}          & {46.63}                   & {86.87}         & {\multirow{2}{*}{predicted}}                 \\ 
                                                   &7 & {arg-LED-base
                                                (fgrain)
                                                + diverse beams}                       & {50.92}                   & {26.06} & {47.74}& {86.87} &   \\ 
                                                \cline{2-8}
                                                   &8 & baseline ranking                     & 51.85                   & 27.31          & 48.61                   & 87.26          & \multirow{2}{*}{oracle}                    \\ 
                                                   &9 & arg-LED-base (fgrain)
                                                   + diverse beams                        & 52.74                   & \textbf{27.93} & 49.50                   & \textbf{87.46}         &                  \\ 
                                                   \hline 
\multirow{4}{*}{\textbf{Our framework}} &10 & {arg-augmented-LED}                                  & {50.52 }                  & {24.82}          & {47.19}                   & {86.85}        & \multirow{2}{*}{{predicted}}                 \\
            &11* & \textbf{arg-augmented-LED + diverse beams} & \textit{{54.13}} & {\textit{27.02}} & {\textit{50.14}} & \textit{{87.38}} &         \\
                                                   \cline{2-8}

                                                &12 &   arg-augmented-LED                                   &  51.96                   & 25.69          & 48.56                   & 87.03         & \multirow{2}{*}{oracle}                 \\
                                                   
                                                   &13& {\textbf{arg-augmented-LED + diverse beams}} & \textbf{54.30}
                                                   & 27.00 & \textbf{50.80} & 87.35& 
\\
                                                   
                                                   \hlinewd{1.5pt}
\end{tabular}
\begin{center}
   \caption{\label{res} Summarization 
 ROUGE (R1, R2, RL) and BertScore (BS) cross-validation results. Best results in each column are {\bf bolded} when obtained with the oracle markers and \textit{italicized} with predicted markers. For  full framework (rows 11/13), * indicates  results are statistically significant in all scores over  best argument-aware baseline (row-3).
 }
\end{center}

\end{table*}
\end{center}

\subsection{Scoring and Reranking Summaries}
We propose a scoring method to rank the candidate summaries based on their capability to capture the main argument points in the input. 
First, we employ a sentence-level argument role classifier to extract sentences with argument roles $\hat{X}_{args}$. 
The predicted sentences are used to construct an extractive summary.  Then, we measure the lexical overlap between a generated candidate summary $\hat{S}$ and the constructed extracted one using \textit{ROUGE-1 F1-score}\footnote{Using R-2 or R-L makes little difference 
(Appendix \ref{rouge_rank_app}).}, to compute a score to each candidate summary that represents its alignment with the legal opinion argument content. Our scoring function $\mu$ can be written as 
$\mu = ROUGE\-1(\hat{X}_{args}, \hat{S})$.



\section{Experiments}

All models use \textit{LED-base} checkpoint as a base model. \textit{LED-base} encodes up to $16k$ tokens, which fits 
our long inputs.
 All experiments use 5-fold cross-validation, with  
 the 4-fold documents split into $90\%$ training and $10\%$ validation; the validation split is used to select the best checkpoint.\footnote{
 Full experimental details 
 can be found in Appendix \ref{exp_setup}.}


%
We compare  all rank-based methods (baseline and proposed) to {\bf abstractive baselines} previously explored in legal opinion summarization: 
\textit{finetune LED-base} (which refers to vanilla model finetuning  using our  dataset), and 
\textit{arg-LED-base}  \cite{elaraby-litman-2022-arglegalsumm} (which  finetunes LED on the {\it dataset  blended with argument markers} that mark the start and the end of each argument role in the input).\footnote{Argument marker details can be found in Appendix \ref{markers}.} 


We also compare our proposed rank-based approach from Section~\ref{sec:models} with \textbf{ranking baselines} that use different input formats or diverse decoding alone. Specifically, we have employed ranking on top of the output of the three LED models outlined in  \citet{elaraby-litman-2022-arglegalsumm} which are trained on distinct argument aware input formats (we refer to this model as "baseline 
ranking"). Additionally, for diverse decoding, we have employed different beam widths within the range of 1 and 5 \footnote { We ran out of memory with BeamWidth > 5.} on top of the model trained on the input with fine-grained markers (arg-LED-fine-grained), which achieved the best abstractive baseline ROUGE results. 

All models utilizing argument markers
employed both {\it oracle} and {\it predicted} conditions during inference time, using human annotations  or argument mining respectively, to produce the 
markers. 
 
\section{Results and Discussion}

Table \ref{res} shows our 
results in terms of
 \textit{ROUGE-score} \cite{lin2004rouge} and \textit{BERTScore} \cite{zhang2019bertscore}, computed using 
 \textit{SummEval} \cite{fabbri-etal-2021-summeval}\footnote{\url{https://github.com/Yale-LILY/SummEval}}.

{\bf Utility of any Ranking}
The ranking-based methods (rows 6-13) consistently outperform the abstractive 
baselines\footnote{See Appendix~\ref{extractive} for extractive baseline results.} (rows 1-5) 
in both predicted and oracle conditions. Also, abstractive baseline results   
(rows 1-5) align with those of \citet{elaraby-litman-2022-arglegalsumm}, where 
leveraging fine-grained markers in the input yields the highest 
scores. 

{\bf Utiliy of Proposed Ranking Framework and its Components}
In the predicted case, our proposed arg-augmented-LED  (row 10) improves over the abstractive baselines (rows 1-3) with ranges $1.5-3.19$ and $1.27-3.07$ in ROUGE-1 and ROUGE-L respectively, while maintaining a limited drop of $0.1$ and $0.01$ in terms of ROUGE-2 and BS respectively. Similarly, compared to our ranking baselines, our proposed model improves over ROUGE-1 and ROUGE-L scores obtained by baseline ranking 
with ranges $0.56-0.73$ while dropping in ROUGE-2 and BS 
by $0.31$ and $0.02$ points respectively. This indicates that incorporating argument information into the source inputs can lead to the generation of effective summary candidates. Our best predicted results were achieved by combining our proposed model with diverse beam decoding (row 11), which combines the strengths of various input formats and multiple beam decoding, resulting in statistically significant improvements over the previously proposed argument-aware abstractive baseline (row 3). 


{\bf Inference with Predicted versus Oracle Argument Roles} 
For the same model,  predicted markers can impact the summarization results. In prior baselines (rows 3 and 5), we observe a drop in ROUGE score with ranges $2.05-2.14$, and  $0.06$ in terms of BS when switching from oracle to predicted markers. This observation is consistent among row 6 and 8; and row 10 and 12. With our proposed arg-augmented-LED and diverse beam decoding, this performance gap is mitigated and reduced to $-0.02 - 0.66$ and $-0.03$ in ROUGE and BS, respectively (rows 11 and 13). We believe this is due to the combination of distinct argumentative formats and diverse decoding, allowing more diverse candidates to be considered in the ranking and enhancing  robustness to noisy predictions during inference.

\section{Conclusion and Future Work}

We proposed a framework for improving the summarization of long legal opinions by combining distinct argument formats of the input with diverse decoding to generate candidate summaries. Our framework selects the summary with the highest lexical overlap with the input's argumentative content. Our results indicate that ranking alone can improve over abstractive baselines. Moreover, combining ranking with our proposed candidate generation method improves results while maintaining robustness to noisy predictions. In future research, we plan to incorporate human expert evaluations to compare automatic metrics with human ratings. Also, we aim to explore the impact of using noisier argument roles during training on a larger corpus by using the predicted markers obtained from our smaller dataset to experiment with the remaining unannotated portion of the CanLII dataset.

\section*{Limitations}
The primary constraints encountered in our research result from our dependence on a single dataset for experimentation and computing resource limitations. Despite these, we postulate that our ranking-based methodology can be utilized for any summarization task that necessitates robust correspondence with a specific structure within the input. To validate this hypothesis, further experimentation is required to assess the generalizability of our technique to alternative datasets and domains. In addition, our limited computational resources prevented us from experimenting with other long document encoder-decoder models such as BigBird and LongT5 \cite{michalopoulos-etal-2022-icdbigbird, guo-etal-2022-longt5} as well as using higher beam widths during decoding. 
Furthermore, the cost and complexity of procuring expert evaluators within the legal domain resulted in using automatic metrics alone.

\section*{Ethical Considerations}
The usage of the generated summary results from legal opinions remains important. Abstractive summarization models have been found to contain hallucinated artifacts that do not come from the source texts \cite{kryscinski-etal-2019-neural, zhao-etal-2020-reducing,kryscinski-etal-2020-evaluating}. While our model incorporated the argument structure of the source article, the generation results may still carry certain levels of non-factual information and need to be utilized with extra care. Similarly, as mentioned in the prior line of works using CanLII \cite{elaraby-litman-2022-arglegalsumm, zhong2022computing}, CanLII has taken
measures to limit the disclosure of defendants’ identities (such as blocking search indexing). Abstractive approaches may cause user information leakage. Thus using the dataset needs to be cautious to avoid impacting those efforts.  

\section*{Acknowledgements}
This material is based upon work supported by the National Science Foundation under Grant No. 2040490 and by Amazon. We would like to thank the members of both the Pitt AI Fairness and Law Project and the Pitt PETAL group, as well as the anonymous reviewers, for valuable comments in improving this work.




\bibliography{anthology,custom}

\begin{thebibliography}{33}
\expandafter\ifx\csname natexlab\endcsname\relax\def\natexlab#1{#1}\fi

\bibitem[{Agarwal et~al.(2022)Agarwal, Xu, and
  Grabmair}]{agarwal2022extractive}
Abhishek Agarwal, Shanshan Xu, and Matthias Grabmair. 2022.
\newblock \href {https://aclanthology.org/2022.findings-emnlp.134} {Extractive
  summarization of legal decisions using multi-task learning and maximal
  marginal relevance}.
\newblock In \emph{Findings of the Association for Computational Linguistics:
  EMNLP 2022}, pages 1857--1872, Abu Dhabi, United Arab Emirates. Association
  for Computational Linguistics.

\bibitem[{Beltagy et~al.(2020)Beltagy, Peters, and
  Cohan}]{beltagy2020longformer}
Iz~Beltagy, Matthew~E Peters, and Arman Cohan. 2020.
\newblock Longformer: The long-document transformer.
\newblock \emph{arXiv preprint arXiv:2004.05150}.

\bibitem[{Devlin et~al.(2019)Devlin, Chang, Lee, and
  Toutanova}]{devlin2019bert}
Jacob Devlin, Ming-Wei Chang, Kenton Lee, and Kristina Toutanova. 2019.
\newblock Bert: Pre-training of deep bidirectional transformers for language
  understanding.
\newblock In \emph{Proceedings of the 2019 Conference of the North American
  Chapter of the Association for Computational Linguistics: Human Language
  Technologies, Volume 1 (Long and Short Papers)}, pages 4171--4186.

\bibitem[{DeYoung et~al.(2021)DeYoung, Beltagy, van Zuylen, Kuehl, and
  Wang}]{deyoung2021ms2}
Jay DeYoung, Iz~Beltagy, Madeleine van Zuylen, Bailey Kuehl, and Lucy Wang.
  2021.
\newblock Msˆ2: Multi-document summarization of medical studies.
\newblock In \emph{Proceedings of the 2021 Conference on Empirical Methods in
  Natural Language Processing}, pages 7494--7513.

\bibitem[{Dong et~al.(2021)Dong, Mircea, and Cheung}]{dong2021discourse}
Yue Dong, Andrei Mircea, and Jackie Chi~Kit Cheung. 2021.
\newblock Discourse-aware unsupervised summarization for long scientific
  documents.
\newblock In \emph{Proceedings of the 16th Conference of the European Chapter
  of the Association for Computational Linguistics: Main Volume}, pages
  1089--1102.

\bibitem[{Elaraby and Litman(2022)}]{elaraby-litman-2022-arglegalsumm}
Mohamed Elaraby and Diane Litman. 2022.
\newblock \href {https://aclanthology.org/2022.coling-1.540}
  {{A}rg{L}egal{S}umm: Improving abstractive summarization of legal documents
  with argument mining}.
\newblock In \emph{Proceedings of the 29th International Conference on
  Computational Linguistics}, pages 6187--6194, Gyeongju, Republic of Korea.
  International Committee on Computational Linguistics.

\bibitem[{Fabbri et~al.(2021{\natexlab{a}})Fabbri, Kry{\'s}ci{\'n}ski, McCann,
  Xiong, Socher, and Radev}]{fabbri-etal-2021-summeval}
Alexander~R. Fabbri, Wojciech Kry{\'s}ci{\'n}ski, Bryan McCann, Caiming Xiong,
  Richard Socher, and Dragomir Radev. 2021{\natexlab{a}}.
\newblock \href {https://doi.org/10.1162/tacl_a_00373} {{S}umm{E}val:
  Re-evaluating summarization evaluation}.
\newblock \emph{Transactions of the Association for Computational Linguistics},
  9:391--409.

\bibitem[{Fabbri et~al.(2021{\natexlab{b}})Fabbri, Rahman, Rizvi, Wang, Li,
  Mehdad, and Radev}]{fabbri2021convosumm}
Alexander~Richard Fabbri, Faiaz Rahman, Imad Rizvi, Borui Wang, Haoran Li,
  Yashar Mehdad, and Dragomir Radev. 2021{\natexlab{b}}.
\newblock Convosumm: Conversation summarization benchmark and improved
  abstractive summarization with argument mining.
\newblock In \emph{Proceedings of the 59th Annual Meeting of the Association
  for Computational Linguistics and the 11th International Joint Conference on
  Natural Language Processing (Volume 1: Long Papers)}, pages 6866--6880.

\bibitem[{Guo et~al.(2022)Guo, Ainslie, Uthus, Ontanon, Ni, Sung, and
  Yang}]{guo-etal-2022-longt5}
Mandy Guo, Joshua Ainslie, David Uthus, Santiago Ontanon, Jianmo Ni, Yun-Hsuan
  Sung, and Yinfei Yang. 2022.
\newblock \href {https://doi.org/10.18653/v1/2022.findings-naacl.55}
  {{L}ong{T}5: {E}fficient text-to-text transformer for long sequences}.
\newblock In \emph{Findings of the Association for Computational Linguistics:
  NAACL 2022}, pages 724--736, Seattle, United States. Association for
  Computational Linguistics.

\bibitem[{Habernal et~al.(2022)Habernal, Faber, Recchia, Bretthauer, Gurevych,
  Burchard et~al.}]{habernal2022mining}
Ivan Habernal, Daniel Faber, Nicola Recchia, Sebastian Bretthauer, Iryna
  Gurevych, Christoph Burchard, et~al. 2022.
\newblock Mining legal arguments in court decisions.
\newblock \emph{arXiv preprint arXiv:2208.06178}.

\bibitem[{Huang et~al.(2020)Huang, Yu, Guo, Yu, and Xian}]{huang2020legal}
Yuxin Huang, Zhengtao Yu, Junjun Guo, Zhiqiang Yu, and Yantuan Xian. 2020.
\newblock Legal public opinion news abstractive summarization by incorporating
  topic information.
\newblock \emph{International Journal of Machine Learning and Cybernetics},
  11(9):2039--2050.

\bibitem[{Kan et~al.(2021)Kan, Chang, and Chuang}]{kan-etal-2021-home}
Tai-Jung Kan, Chia-Hui Chang, and Hsiu-Min Chuang. 2021.
\newblock \href {https://aclanthology.org/2021.rocling-1.24} {Home appliance
  review research via adversarial reptile}.
\newblock In \emph{Proceedings of the 33rd Conference on Computational
  Linguistics and Speech Processing (ROCLING 2021)}, pages 183--191, Taoyuan,
  Taiwan. The Association for Computational Linguistics and Chinese Language
  Processing (ACLCLP).

\bibitem[{Kanapala et~al.(2019)Kanapala, Pal, and Pamula}]{kanapala2019text}
Ambedkar Kanapala, Sukomal Pal, and Rajendra Pamula. 2019.
\newblock Text summarization from legal documents: a survey.
\newblock \emph{Artificial Intelligence Review}, 51(3):371--402.

\bibitem[{Khalifa et~al.(2021)Khalifa, Ballesteros, and
  Mckeown}]{khalifa2021bag}
Muhammad Khalifa, Miguel Ballesteros, and Kathleen Mckeown. 2021.
\newblock A bag of tricks for dialogue summarization.
\newblock In \emph{Proceedings of the 2021 Conference on Empirical Methods in
  Natural Language Processing}, pages 8014--8022.

\bibitem[{Kingma and Ba(2015)}]{kingma2014adam}
Diederik~P. Kingma and Jimmy Ba. 2015.
\newblock \href {http://arxiv.org/abs/1412.6980} {Adam: {A} method for
  stochastic optimization}.
\newblock In \emph{3rd International Conference on Learning Representations,
  {ICLR} 2015, San Diego, CA, USA, May 7-9, 2015, Conference Track
  Proceedings}.

\bibitem[{Kryscinski et~al.(2019)Kryscinski, Keskar, McCann, Xiong, and
  Socher}]{kryscinski-etal-2019-neural}
Wojciech Kryscinski, Nitish~Shirish Keskar, Bryan McCann, Caiming Xiong, and
  Richard Socher. 2019.
\newblock \href {https://doi.org/10.18653/v1/D19-1051} {Neural text
  summarization: A critical evaluation}.
\newblock In \emph{Proceedings of the 2019 Conference on Empirical Methods in
  Natural Language Processing and the 9th International Joint Conference on
  Natural Language Processing (EMNLP-IJCNLP)}, pages 540--551, Hong Kong,
  China. Association for Computational Linguistics.

\bibitem[{Kryscinski et~al.(2020)Kryscinski, McCann, Xiong, and
  Socher}]{kryscinski-etal-2020-evaluating}
Wojciech Kryscinski, Bryan McCann, Caiming Xiong, and Richard Socher. 2020.
\newblock \href {https://doi.org/10.18653/v1/2020.emnlp-main.750} {Evaluating
  the factual consistency of abstractive text summarization}.
\newblock In \emph{Proceedings of the 2020 Conference on Empirical Methods in
  Natural Language Processing (EMNLP)}, pages 9332--9346, Online. Association
  for Computational Linguistics.

\bibitem[{Lin(2004)}]{lin2004rouge}
Chin-Yew Lin. 2004.
\newblock Rouge: A package for automatic evaluation of summaries.
\newblock In \emph{Text summarization branches out}, pages 74--81.

\bibitem[{Liu et~al.(2019)Liu, Ott, Goyal, Du, Joshi, Chen, Levy, Lewis,
  Zettlemoyer, and Stoyanov}]{liu2019roberta}
Yinhan Liu, Myle Ott, Naman Goyal, Jingfei Du, Mandar Joshi, Danqi Chen, Omer
  Levy, Mike Lewis, Luke Zettlemoyer, and Veselin Stoyanov. 2019.
\newblock Roberta: A robustly optimized bert pretraining approach.
\newblock \emph{arXiv preprint arXiv:1907.11692}.

\bibitem[{Liu and Liu(2021)}]{liu2021simcls}
Yixin Liu and Pengfei Liu. 2021.
\newblock Simcls: A simple framework for contrastive learning of abstractive
  summarization.
\newblock In \emph{Proceedings of the 59th Annual Meeting of the Association
  for Computational Linguistics and the 11th International Joint Conference on
  Natural Language Processing (Volume 2: Short Papers)}, pages 1065--1072.

\bibitem[{Liu et~al.(2022)Liu, Liu, Radev, and Neubig}]{liu2022brio}
Yixin Liu, Pengfei Liu, Dragomir Radev, and Graham Neubig. 2022.
\newblock Brio: Bringing order to abstractive summarization.
\newblock In \emph{Proceedings of the 60th Annual Meeting of the Association
  for Computational Linguistics (Volume 1: Long Papers)}, pages 2890--2903.

\bibitem[{Michalopoulos et~al.(2022)Michalopoulos, Malyska, Sahar, Wong, and
  Chen}]{michalopoulos-etal-2022-icdbigbird}
George Michalopoulos, Michal Malyska, Nicola Sahar, Alexander Wong, and Helen
  Chen. 2022.
\newblock \href {https://doi.org/10.18653/v1/2022.bionlp-1.32} {{ICDB}ig{B}ird:
  A contextual embedding model for {ICD} code classification}.
\newblock In \emph{Proceedings of the 21st Workshop on Biomedical Language
  Processing}, pages 330--336, Dublin, Ireland. Association for Computational
  Linguistics.

\bibitem[{Moro and Ragazzi(2022)}]{moro2022semantic}
Gianluca Moro and Luca Ragazzi. 2022.
\newblock Semantic self-segmentation for abstractive summarization of long
  legal documents in low-resource regimes.
\newblock In \emph{Proceedings of the Thirty-Six AAAI Conference on Artificial
  Intelligence, Virtual}, volume~22.

\bibitem[{Oved and Levy(2021)}]{oved2021pass}
Nadav Oved and Ran Levy. 2021.
\newblock Pass: Perturb-and-select summarizer for product reviews.
\newblock In \emph{Proceedings of the 59th Annual Meeting of the Association
  for Computational Linguistics and the 11th International Joint Conference on
  Natural Language Processing (Volume 1: Long Papers)}, pages 351--365.

\bibitem[{Ravaut et~al.(2022)Ravaut, Joty, and Chen}]{ravaut2022summareranker}
Mathieu Ravaut, Shafiq Joty, and Nancy Chen. 2022.
\newblock Summareranker: A multi-task mixture-of-experts re-ranking framework
  for abstractive summarization.
\newblock In \emph{Proceedings of the 60th Annual Meeting of the Association
  for Computational Linguistics (Volume 1: Long Papers)}, pages 4504--4524.

\bibitem[{Wolf et~al.(2020)Wolf, Debut, Sanh, Chaumond, Delangue, Moi, Cistac,
  Rault, Louf, Funtowicz, Davison, Shleifer, von Platen, Ma, Jernite, Plu, Xu,
  Le~Scao, Gugger, Drame, Lhoest, and Rush}]{wolf2019huggingface}
Thomas Wolf, Lysandre Debut, Victor Sanh, Julien Chaumond, Clement Delangue,
  Anthony Moi, Pierric Cistac, Tim Rault, Remi Louf, Morgan Funtowicz, Joe
  Davison, Sam Shleifer, Patrick von Platen, Clara Ma, Yacine Jernite, Julien
  Plu, Canwen Xu, Teven Le~Scao, Sylvain Gugger, Mariama Drame, Quentin Lhoest,
  and Alexander Rush. 2020.
\newblock \href {https://doi.org/10.18653/v1/2020.emnlp-demos.6} {Transformers:
  State-of-the-art natural language processing}.
\newblock In \emph{Proceedings of the 2020 Conference on Empirical Methods in
  Natural Language Processing: System Demonstrations}, pages 38--45, Online.
  Association for Computational Linguistics.

\bibitem[{Xu and Ashley(2022)}]{xu2022multi}
Huihui Xu and Kevin~D. Ashley. 2022.
\newblock Multi-granularity argument mining in legal texts.
\newblock In \emph{International Conference on Legal Knowledge and Information
  Systems}.

\bibitem[{Xu et~al.(2021)Xu, Savelka, and Ashley}]{xu2021toward}
Huihui Xu, Jaromir Savelka, and Kevin~D Ashley. 2021.
\newblock Toward summarizing case decisions via extracting argument issues,
  reasons, and conclusions.
\newblock In \emph{Proceedings of the eighteenth international conference on
  artificial intelligence and law}, pages 250--254.

\bibitem[{Zhang et~al.(2019)Zhang, Kishore, Wu, Weinberger, and
  Artzi}]{zhang2019bertscore}
Tianyi Zhang, Varsha Kishore, Felix Wu, Kilian~Q Weinberger, and Yoav Artzi.
  2019.
\newblock Bertscore: Evaluating text generation with bert.
\newblock In \emph{International Conference on Learning Representations}.

\bibitem[{Zhao et~al.(2020)Zhao, Cohen, and Webber}]{zhao-etal-2020-reducing}
Zheng Zhao, Shay~B. Cohen, and Bonnie Webber. 2020.
\newblock \href {https://doi.org/10.18653/v1/2020.findings-emnlp.203} {Reducing
  quantity hallucinations in abstractive summarization}.
\newblock In \emph{Findings of the Association for Computational Linguistics:
  EMNLP 2020}, pages 2237--2249, Online. Association for Computational
  Linguistics.

\bibitem[{Zheng and Lapata(2019)}]{zheng2019sentence}
Hao Zheng and Mirella Lapata. 2019.
\newblock Sentence centrality revisited for unsupervised summarization.
\newblock In \emph{Proceedings of the 57th Annual Meeting of the Association
  for Computational Linguistics}, pages 6236--6247.

\bibitem[{Zheng et~al.(2021)Zheng, Guha, Anderson, Henderson, and
  Ho}]{zhengguha2021}
Lucia Zheng, Neel Guha, Brandon~R. Anderson, Peter Henderson, and Daniel~E. Ho.
  2021.
\newblock \href {http://arxiv.org/abs/2104.08671} {When does pretraining help?
  assessing self-supervised learning for law and the casehold dataset}.
\newblock In \emph{Proceedings of the 18th International Conference on
  Artificial Intelligence and Law}. Association for Computing Machinery.

\bibitem[{Zhong and Litman(2022)}]{zhong2022computing}
Yang Zhong and Diane Litman. 2022.
\newblock \href {https://aclanthology.org/2022.nllp-1.30} {Computing and
  exploiting document structure to improve unsupervised extractive
  summarization of legal case decisions}.
\newblock In \emph{Proceedings of the Natural Legal Language Processing
  Workshop 2022}, pages 322--337, Abu Dhabi, United Arab Emirates (Hybrid).
  Association for Computational Linguistics.

\end{thebibliography}
\bibliographystyle{acl_natbib}

\appendix

\section{Argument Role Labeling in CanLII Cases}
\label{irc_des}

\begin{figure*}[h]
\small
\begin{center}

 \includegraphics[width=16cm,height=8cm]{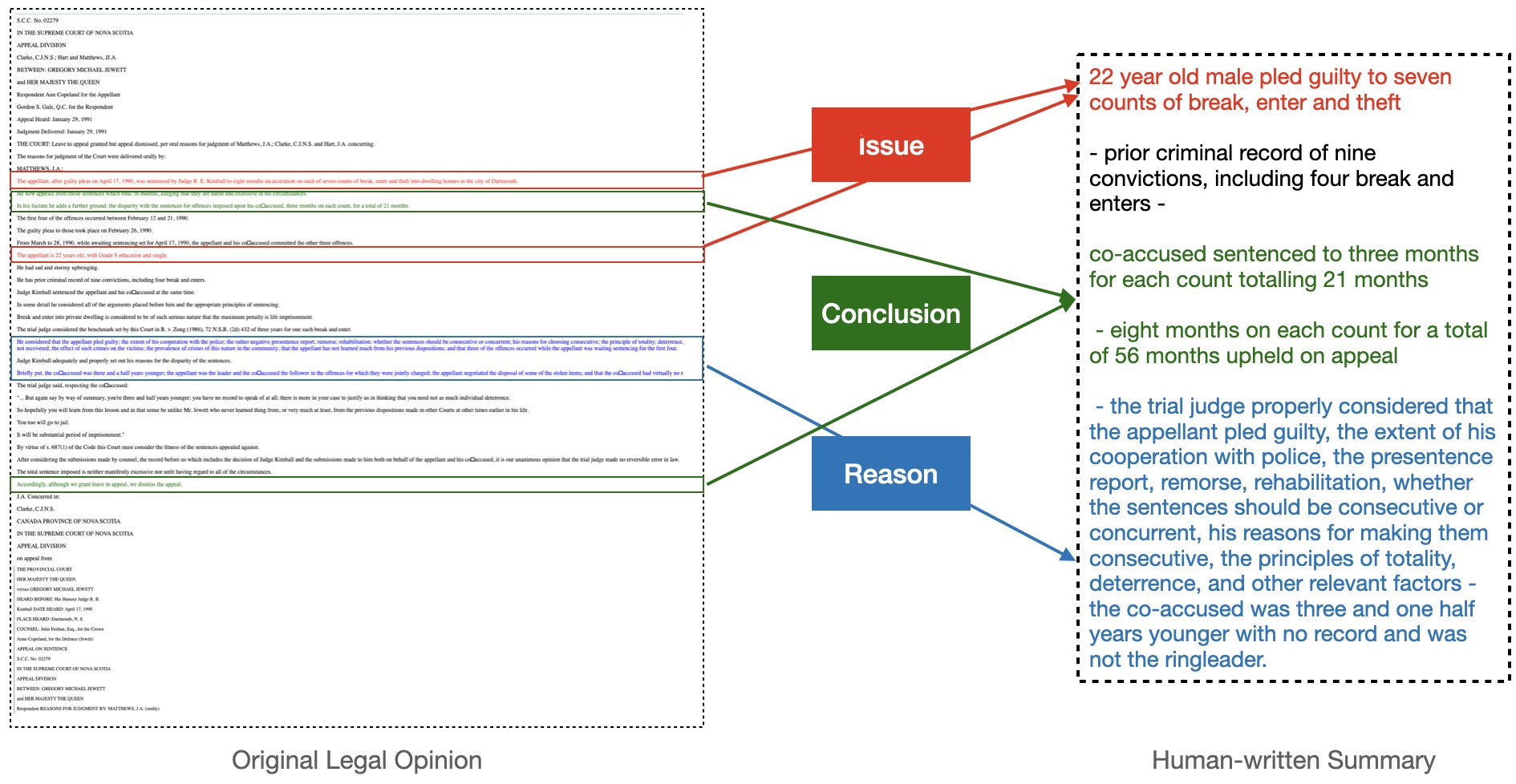}
 \caption{\label{irc_example} An example of the annotated Issue, Reason, and Conclusion sentences in the CanLII dataset’s legal opinion and
summary pair (ID: a\_1991canlii2497).}
 \end{center}
\end{figure*}

The concept of argument roles, specifically issues, reasons, and conclusions, is of paramount importance in legal case summarization. An illustration, presented in Figure \ref{irc_example}, demonstrates the annotation of these roles in the input text of a legal opinion and its associated summary. This example shows that the issues, reasons, and conclusions can effectively encapsulate the critical points of discussion within the court, the ultimate decision reached, and the rationale for said decision.

\section{Experimental Setup and Hyper-parameters}
\label{exp_setup}
\textbf{LED experiments}
For all of our LED-base experiments, we use the LED-base implementation by the \textit{HuggingFace Library} \cite{wolf2019huggingface}.  
We finetune the LED-base model for $10$ epochs. We select our best model based on the $ROUGE-2$ score on the validation set. We rely on the Adam optimizer \cite{kingma2014adam} with an initial learning rate of $2e-5$ to update the LED-base weights. We also employ an early stopping with $3$ epoch patience to avoid overfitting during training. 

\textbf{Argument Role Classification}
 Our argument role classifier leverages a finetuned \textit{legalBERT} \cite{zhengguha2021} model due to its superiority to other contextualized embeddings-based models like BERT \cite{devlin2019bert} and ROBERTa \cite{liu2019roberta} as shown in \citet{elaraby-litman-2022-arglegalsumm, xu2021toward}. We utilized the same training setting and hyperparameters described in \citet{elaraby-litman-2022-arglegalsumm} to train the \textit{5-fold} cross-validation sentence level argument classifiers used in our experiments. 
 \footnote{Classifier code is available at \url{https://github.com/EngSalem/arglegalsumm/tree/master/src/argument_classification}}

\begin{table}[t]
\small
\begin{center}
\begin{tabular}{p{0.1\textwidth}|p{0.29\textwidth}}
\hlinewd{1.5pt}
                    &
                    
                    {\textbf{Example of using argument markers}}\\ \hlinewd{1.5pt}                                                                                                                    
\textbf{Original}            & The plaintiful should have taken more appropriate measures to avoid the accident.                                                                \\
\hline
\textbf{Binary Markers}      & \textbf{\texttt{<IRC>}} The plaintiful should have taken more appropriate measures to avoid the accident. \textbf{\texttt{</IRC>}} .      \\
\hline
\textbf{Fine-grained Markers} & \textbf{\texttt{<Reason>}} The plaintiful should have taken more appropriate measures to avoid the accident. \textbf{\texttt{</Reason>}.}\\
\hlinewd{1.5pt}
\end{tabular}

\caption{Example of using argumentative marker tokens}
\label{arg_example}
\end{center}

\end{table}

\section{Argumentative Markers}
\label{markers}


In abstractive summarization, special markers can indicate the most important parts of a text that ground the summary \cite{khalifa2021bag, deyoung2021ms2}. These markers can be added to the text by a human annotator, or they can be generated automatically by a  model. These markers can take many forms, such as highlighting certain words or phrases or adding special tags to certain sentences. A summarization model can use them to identify the key parts of the text that should be included in the summary while also considering the overall structure and coherence of the text. This can help to improve the accuracy and effectiveness of the summarization process, especially when the text is long or complex. In this work, we use marker sets proposed by \citet{elaraby-litman-2022-arglegalsumm} to distinguish between argumentative and non-argumentative sentences.

\textbf{Binary markers}
The binary markers aim to distinguish argumentative and non-argumentative sentences regardless of the type of the argument role (i.e,  issues, reasons, or conclusions). In our work, we used the markers \texttt{<IRC>,</IRC>} to highlight the start and end of each argumentative sentence. 

\textbf{Fine-grained markers}
 We also used the markers designated to distinguish between each argument role type by using the markers \texttt{<Issue>,</Issue>, <Reason>, </Reason>, <Conclusion>, </Conclusion>}.

Table \ref{arg_example} shows an example of using different argumentative markers to highlight the start and end of a "Reason" sentence. 



\begin{center}
    
\begin{table*}[b]
\small
\renewcommand{\arraystretch}{1.15}
\begin{tabular}{p{0.17\textwidth}cp{0.31\textwidth}lllll}

\hlinewd{1.5pt}
\textbf{Experiments}          &    ID                          & \textbf{Model }                                              & \textbf{R-1 }                & \textbf{R-2}        & \textbf{R-L}                 & \textbf{BS}  & \textbf{src. marker} \\ \hlinewd{1.5pt}

\multirow{3}{*}{\textbf{Extractive baselines}} 
  & 1 & sentence-level legalBERT & 49.66 & 28.42 & 46.72  & 86.54
                                             & -\\
                                             & 2
                                             & HipoRank                                            & 41.24                   & 17.19          & 38.54                   & 81.67      &           -           \\  
                                                 &
                                                 3
                                                 & HipoRank rewighted                                  & 42.88                   & 18.03          & 39.99                  & 84.11     &            -          \\
                                                &
                                                4
                                                 & Extractive BERT                                  & 43.053                     &     17.75          &   39.99                    & 84.15         & -
                                                 
                                                \\ \hline

\multirow{5}{*}{\textbf{Abstractive baselines}}   &5   & finetune LED-base  & 47.33                   & 22.80   & 44.12 & 86.43  & -  \\  
    \cline{2-8} 
    &6 & \textit{arg-LED-base (binary markers)}                       & \textit{48.85}                   & \textit{24.74}          & \textit{45.82}                   & \textit{86.79}          & \multirow{2}{*}{\textit{predicted }}                 \\  
                                                 &7 &  \textit{arg-LED-base (fine-grained markers)}                 & \textit{49.02}                   & \textit{24.92}          & \textit{45.92}                   & \textit{86.86}         &                 \\ 
                                                    \cline{2-8}
   
     &8 & arg-LED-base (binary markers)                      & 50.64                   & 26.62          & 47.48                   & 86.90          & \multirow{2}{*}{oracle}            \\  
                                                  &9 & arg-LED-base ( fine-grained markers)                 & 51.07                   & 27.06          & 48.01                   & 86.92         &                 \\

                                                   \hline
\multirow{4}{*}{\textbf{Ranking baselines}}            &10 & \textit{baseline ranking}                           & \textit{49.79}                   & \textit{25.13}          & \textit{46.63}                   & \textit{86.87}         & \textit{\multirow{2}{*}{predicted}}                 \\ 
                                                   &11 & \textit{arg-LED-base + diverse beams }                       & \textit{50.92}                   & \textit{26.06} & \textit{47.74}                   & \textit{86.87}          &                 \\ 
                                                   \cline{2-8}

                                                   &12 & baseline ranking                     & 51.85                   & 27.31          & 48.61                   & 87.26          & \multirow{2}{*}{oracle}                    \\ 
                                                   &13 & arg-LED-base + diverse beams                        & 52.74                   & \textbf{27.93} & 49.50                   & \textbf{87.46}         &                  \\ 
                                                   \hline 
\multirow{4}{*}{\textbf{Our framework}} &14 & \textit{arg-augmented-LED}                                  & \textit{50.52 }                  & \textit{24.82}          & \textit{47.19}                   & \textit{86.85}        & \multirow{2}{*}{\textit{predicted}}                 \\
            &15 & \textit{\textbf{arg-augmented-LED + diverse beams}} & \textit{\textbf{54.13}} & \textit{\textbf{27.02}} & \textit{\textbf{50.14}} & \textbf{\textit{87.38}} & \        \\
                                                   \cline{2-8}

                                                &16 &   arg-augmented-LED                                   &  51.96                   & 25.69          & 48.56                   & 87.03         & \multirow{2}{*}{oracle}                 \\
                                                   
                                                   &17& \textbf{arg-augmented-LED + diverse beams} & \textbf{54.30}
                                                   & 27.00 & \textbf{50.80} & 87.35& 
\\
                                                   
                                                   \hlinewd{1.5pt}
\end{tabular}
\caption{Full Extractive and Abstractive Results \label{ext_res}}
\end{table*}
\end{center}

\section{Extractive Baselines}
\label{extractive}

 In addition to the abstractive baselines, we compare our methods to graph-based unsupervised extractive baselines built on top of \textit{HipoRank} \cite{dong2021discourse} and extractive baselines based on \textit{Extractive-BERT} \cite{zheng2019sentence}, which were leveraged before on the same dataset \cite{zhong2022computing}. Table \ref{ext_res} shows our abstractive summarization results compared to the extractive baselines in cross-validation settings. Our ranking-based methods show consistent improvement over both the extractive and the abstractive baselines.

\section{ROUGE based ranking results}
\label{rouge_rank_app}

Table \ref{rouge_rank} shows a comparison between the usage of ROUGE-1, ROUGE-2, and ROUGE-L as potential ranking criteria to select the summary that aligns with the predicted argumentative content outlined in the input legal opinion. While there is no substantial differences between results with each ROUGE metric, ROUGE-L seems to have marginally lower scores compared to ROUGE-1, and ROUGE-2.  
\begin{table*}[h]
\begin{center}
\small
\begin{tabular}{llllll}
\hlinewd{1.5pt}
\textbf{Ranking metric}           & \textbf{Model}                            & \textbf{R-1}   & \textbf{R-2}   & \textbf{R-L}   & \textbf{BS}    \\
\hlinewd{1.5pt}
\multirow{4}{*}{\textbf{ROUGE-1}} & baseline ranking                     & 49.79 & 25.14 & 46.63 & 86.87 \\
                         & arg-LED + diverse beams          & 50.92 & 26.06 & 47.74 & 86.87 \\
                         & arg-augmented-LED                 & 50.52                   & 24.82          & 47.19                   & 86.85    \\
                         & arg-augmented-LED + diverse beams &  54.13 & 27.02 & 50.14 & 87.38 \\ \hline
\multirow{4}{*}{\textbf{ROUGE-2}} & baseline ranking                     & 49.39 & 25.27 & 46.30 & 86.88 \\
                         & arg-LED + diverse beams          & 50.35 & 26.24 & 47.23 & 87.15 \\
                         & arg-augmented-LED                 & 49.46 & 24.16 & 46.19 & 86.71 \\
                         & arg-augmented-LED + diverse beams &   54.00    &  27.82     &  50.08     &  87.42     \\ \hline
\multirow{4}{*}{\textbf{ROUGE-L}} & baseline ranking                    & 49.12 & 24.97 & 46.01 & 86.84 \\
                         & arg-LED + diverse beams          & 49.84 & 25.66 & 46.72 & 87.07 \\
                         & arg-augmented-LED                 & 49.16 & 23.87 & 45.87 & 86.67 \\
                         & arg-augmented-LED + diverse beams & 53.34       &  26.88     &   49.98    &   87.39
\\
\hlinewd{1.5pt}                         
\end{tabular}
\caption{R1, R2, RL ranking scores with predicted argumentative markers \label{rouge_rank}}
    
\end{center}
\end{table*}

\end{document}